\documentclass{article}

\PassOptionsToPackage{numbers, compress}{natbib}

\usepackage[final]{AdvML_Frontiers_2024}

\usepackage[utf8]{inputenc} 
\usepackage[T1]{fontenc}    
\usepackage{hyperref}       
\usepackage{url}            
\usepackage{booktabs}       
\usepackage{amsfonts}       
\usepackage{nicefrac}       
\usepackage{microtype}      
\usepackage{xcolor}         

\usepackage{multirow}
\usepackage{siunitx}
\usepackage{pgfplots, tikz}
\usepackage{amsmath}
\usepackage{natbib}
\usepackage{float}

\title{Learning From Convolution-based Unlearnable Datasets}

\author{Dohyun Kim \quad Pedro Sandoval-Segura \\
  University of Maryland, College Park\\
  \texttt{dkim5124@terpmail.umd.edu} \quad \texttt{psando@umd.edu}
}

\begin{document}

\maketitle

\begin{abstract}
The construction of large datasets for deep learning has raised concerns regarding unauthorized use of online data, leading to increased interest in protecting data from third-parties who want to use it for training. 
The Convolution-based Unlearnable DAtaset (CUDA) method aims to make data unlearnable by applying class-wise blurs to every image in the dataset so that neural networks learn relations between blur kernels and labels, as opposed to informative features for classifying clean data.
In this work, we evaluate whether CUDA data remains unlearnable after image sharpening and frequency filtering, finding that this combination of simple transforms improves the utility of CUDA data for training. In particular, we observe a substantial increase in test accuracy over adversarial training for models trained with CUDA unlearnable data from CIFAR-10, CIFAR-100, and ImageNet-100. In training models to high accuracy using unlearnable data, we underscore the need for ongoing refinement in data poisoning techniques to ensure data privacy. 
Our method opens new avenues for enhancing the robustness of unlearnable datasets by highlighting that simple methods such as sharpening and frequency filtering are capable of breaking convolution-based unlearnable datasets. \footnote{Our code is available at \url{https://github.com/aseriesof-tubes/RSK}}
\end{abstract}

\begin{table}[h]
\centering
\begin{sc}
\begin{tabular}{lcccc}
\toprule
\multirow{2}{*}{} & \multicolumn{1}{l}{\multirow{2}{*}{Clean Data}} & \multicolumn{1}{l}{\multirow{2}{*}{CUDA Data}} & \multicolumn{2}{c}{CUDA Data with} \\
                  & \multicolumn{1}{l}{}                       & \multicolumn{1}{l}{}                      & AT     & RSK+FF (Ours)   \\ \midrule
CIFAR-10              & 94.34     & 23.25       & 44.40        & \textbf{78.53}         \\
CIFAR-100             & 74.41     & 17.33       & 34.34        & \textbf{53.33}         \\
ImageNet-100            & 81.76     & 2.68        & 38.68       & \textbf{43.08}   

\end{tabular}
\end{sc}
\vspace{0.2cm}
\caption{\textbf{Compared to common defenses like Adversarial Training (AT), our method applied to CUDA data can train a RN-18 to higher test accuracy on CIFAR-10, CIFAR-100, and ImageNet-100}. Our method uses Random Sharpening Kernels and Frequency Filtering (RSK+FF) to alter poisoned CUDA images during training. On CIFAR-10, CIFAR-100, and ImageNet-100, we achieve 55\%, 36\%, and 40\% improvements, respectively, over training with CUDA data.}
\label{table:teaser-table}
\end{table}
\section{Introduction}

Widespread data scraping has raised concerns about privacy and securing important data away from malicious actors. \citet{desai2021virtex} mention the need and goal to scale to training on hundreds of millions of images, which is a daunting task without the ability to scrape the web. Additionally, the usage of private data may inadvertently lead to the susceptibility of inference attacks \cite{nasr2018comprehensive}. While the intentions of scrapers may not be malicious in nature, concerns of privacy remain. Poisoning can be used as a countermeasure against web scraping, which negatively affects the performance of a model that is trained on said dataset \cite{carlini2023poisoning}. Another kind of poisoning attack, backdoor poisoning \cite{chen2017targeted}, poses a threat to learning systems by causing targeted misclassifications. 

To prevent the unauthorized scraping of data on the web, \textit{unlearnable datasets} have been introduced as a way to render images unusable. Unlearnable dataset methods seek to find out the best way to imperceptibly modify a training dataset so that a deep neural network (DNN) trained on the resulting data exhibits the largest generalization gap. The idea is that if data can be rendered useless, third parties are disincentivized from using the modified data. 

Two types of unlearnable datasets exist: bounded and unbounded. Bounded methods \cite{huang2021unlearnable,fowl2021adversarialpoison, sandoval-segura2022autoregressive, fu2022robust} attempt to make the perturbation as imperceptible to humans, while unbounded methods \cite{Sadasivan_2023_CVPR, wu2023onepixel} are more easily perceptible. Both types of unlearnable datasets face challenges. Existing works that explore bounded unlearnable methods have shown that Adversarial Training (AT) \cite{huang2021unlearnable, fowl2021adversarialpoison, yuan21ntga} and JPEG compression \cite{liu2023image, das2017keeping} can increase the learnability of unlearnable data.



We propose a method for learning from unbounded unlearnable datasets like the Convolution-based Unlearnable DAtaset (CUDA). As bounded methods show weakness whether it be by AT or other transforms, \citet{Sadasivan_2023_CVPR} argue that unbounded unlearnable methods are valid because 1) they do not obscure the semantics of the dataset, and 2) their class-wise blurring makes the model learn the relationship between the class-wise filters and their corresponding labels.

We demonstrate that it is possible to learn from CUDA datasets \cite{Sadasivan_2023_CVPR}, despite their unbounded perturbations. We achieve a high test accuracy by sharpening the images from the CUDA dataset using our sharpening kernels followed by frequency filtering, achieving a 55\%, 36\%, and 40\% improvement over CUDA standard training for CIFAR-10, CIFAR-100, and ImageNet-100, respectively. Our results suggest using random sharpening kernels should be a baseline against convolution-based perturbations.

\begin{figure}[t]
    \centering
    \resizebox{10cm}{!}{
        \includegraphics{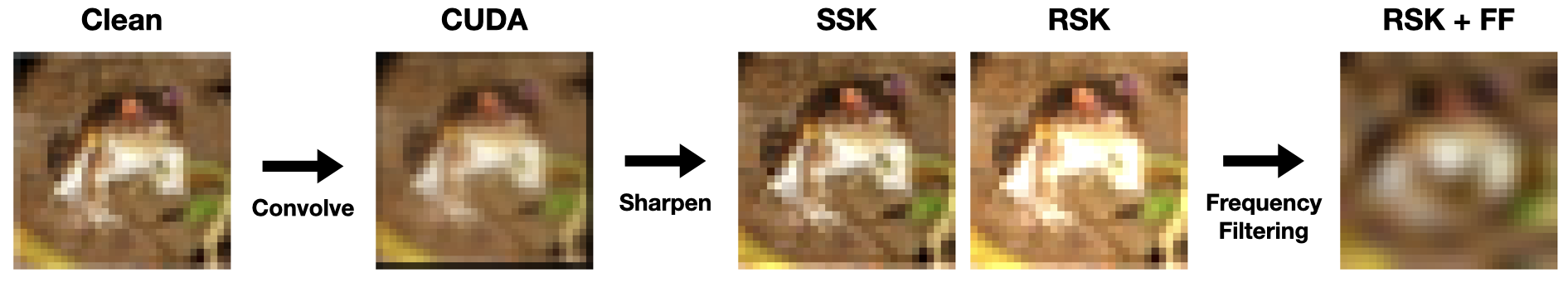}
    }
    \caption{\textbf{Sharpening and Frequency Filtering of a CIFAR-10 Image}. We analyze the effect of standard and random sharpening kernels, both with a center value of 2.5.~We find that the randomized sharpening kernel, denoted RSK, ensures images of the same class are sharpened differently. After sharpening, we decompose the image into spatial frequencies using DCT and filter out high frequencies (see Section~\ref{subsection:frequency-filtering}).}
    \label{fig:transform-flow}
\end{figure}

\section{Related Work}

Unlearnable dataset methods make small perturbations to the whole training dataset so that DNNs trained on the modified data result in poor test accuracy \citep{huang2021unlearnable, fowl2021adversarialpoison, yu2022availability}. Many types of perturbations have been explored: error-minimizing noise \cite{huang2021unlearnable, fu2022robust}, error-maximizing noise \cite{fowl2021adversarialpoison}, autoregressive noise \cite{sandoval-segura2022autoregressive}, low-frequency noise \cite{sandoval2022poisons}, and more. The modified datasets are typically constrained to ``look like'' the original dataset by requiring that the difference between each original image and its modified image has an $\ell_p$ norm bounded by some $\epsilon$. For example, a number of works bound perturbations using an $\ell_\infty$-norm \cite{huang2021unlearnable, fowl2021adversarialpoison, fu2022robust}, whereas other works use an $\ell_2$ norm \cite{yu2022availability,sandoval-segura2022autoregressive}. This bound is also known as an ``imperceptibility constraint,'' because it ensures that image modifications are not easily visible.


But there has been a growing movement allowing for unbounded perturbations \cite{wu2023onepixel, Sadasivan_2023_CVPR}. Unbounded perturbations are naturally more effective at inducing a generalization gap in DNNs due, in part, to the fact that perturbations are larger and the resulting images are noticeably modified. For example, the One-Pixel Shortcut (OPS) dataset \cite{wu2023onepixel} sets a single pixel that induces the largest gap from the mean pixel value of a class in a class-wise manner (all images of a class are modified in the same way). The Convolution-Based Unlearnable DAtaset (CUDA) dataset \cite{Sadasivan_2023_CVPR} also contains unbounded perturbations, where images are blurred class-wise. For example, on CIFAR-10, ten random blur kernels are generated and used to blur training images. This causes trained networks to learn relations between blur kernel and labels, \textit{i.e.} a learning shortcut \cite{geirhos2020shortcut, yu2022availability} which is not informative at test-time. Nevertheless, \citet{wu2023onepixel} and \citet{Sadasivan_2023_CVPR} argue that as long as the semantic information in an image is preserved, unbounded perturbations should be explored. After all, both works demonstrate that unbounded perturbations are more resistant to adversarial training and augmentations. CUDA is also effective against randomized smoothing \cite{cohen2019certified}, as well as Deconvolution-based adversarial training. This observation aligns with the data in Table \ref{table:avg-perturbation-size}, which shows that the \(\ell_p\) norms of CUDA and OPS are substantially higher than their counterparts. Notably, CUDA exhibits a four to six-fold increase in its \(\ell_2\) norm compared to alternative methods. 

\begin{figure*}[t]
    \centering
    \resizebox{9cm}{!}{
        \includegraphics{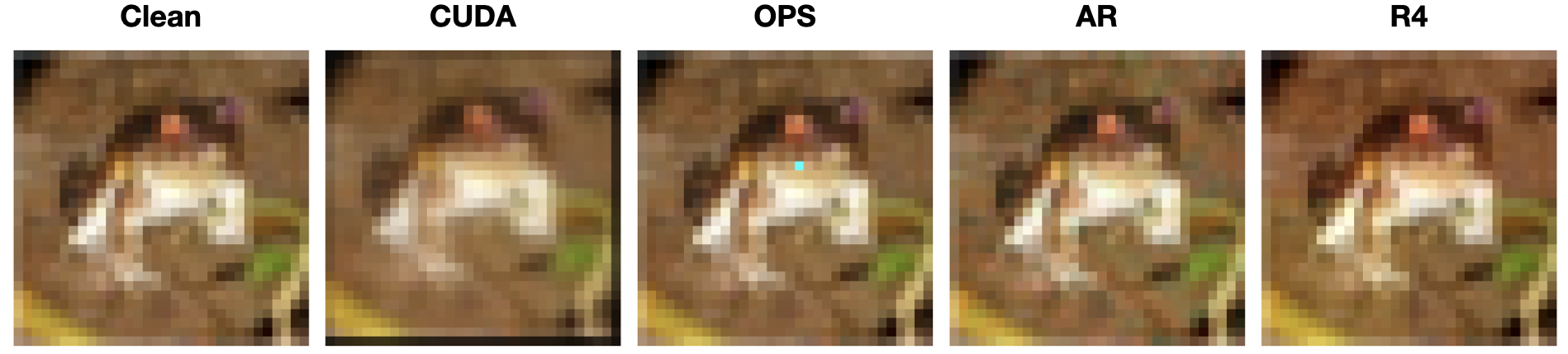}
    }
    \caption{\textbf{Comparison of the same image from different unlearnable datasets.} OPS \cite{wu2023onepixel} perturbs by adding noise to a singular pixel. AR \cite{sandoval-segura2022autoregressive} generates perturbations using a sliding window approach. R4 \cite{sandoval2022poisons} is a grid-like additive perturbation. CUDA and OPS, being unbounded methods, exhibit a noticeable perturbation, compared to AR and R4 which have perturbations bounded by an \(\ell_p\)-norm.}
    \label{fig:ud-comparison}
\end{figure*}

The unbounded nature of CUDA, which causes a visible blur (see Figure~\ref{fig:ud-comparison}), naturally makes it harder for models to learn, as the data is further away from the original training distribution. Because it remains unknown whether recovering the blur kernels from CUDA training data is feasible, CUDA could appear like a reasonable approach to data security -- a data owner could use the CUDA method to protect their data from being trained on. However, in this work, we demonstrate that random sharpening kernels and frequency filtering can make CUDA data learnable again. 

\begin{table}[]
\centering
\begin{sc}
\begin{tabular}{llllllll}
\toprule

              & CUDA  & EM \cite{huang2021unlearnable} & TAP \cite{fowl2021adversarialpoison} & AR \cite{sandoval-segura2022autoregressive} & REM \cite{fu2022robust} & OPS \cite{wu2023onepixel} & R4 \cite{sandoval2022poisons}\\ \midrule
$\ell_2$      & \textbf{6.63} & 1.49 & 1.45 & 0.99 & 1.32 & 1.07 & \underline{1.72}\\
$\ell_\infty$ & \underline{0.54} & 0.03 & 0.03 & 0.09 & 0.03 & \textbf{0.68} & 0.03
\end{tabular}
\end{sc}
\vspace{0.2cm}
\caption{\textbf{Size of an average perturbation from different unlearnable datasets.} For each training image, we compute the norm of the difference between the modified image and the clean image. We average over all training images in CIFAR-10. CUDA perturbations have an $\ell_2$-norm that is at least $3.8\times$ larger than other datasets. OPS is the only other dataset containing unbounded perturbations. We \textbf{bold} the largest norm and \underline{underline} the second-largest norm.}
\label{table:avg-perturbation-size}
\end{table}


 
While some theory can explain why CUDA is effective on data sampled from a Gaussian mixture model~\cite{Sadasivan_2023_CVPR}, our findings on standard datasets like CIFAR-10, CIFAR-100, and ImageNet-100 (a subset of the first 100 classes) demonstrate that classifiers trained on CUDA data can actually still generalize. There have been previous works that attempt to counter unlearnable datasets. Notably, Image Shortcut Squeezing (ISS)~\cite{liu2023image} uses Grayscale and JPEG transforms to counter low-frequency and high-frequency perturbations, respectively. However, the use of JPEG is hard to reason about because of the way JPEG removes information. The JPEG algorithm divides an image into 8x8 blocks then removes the high frequency information from each one independently. This means that it not only removes high frequencies, but the high frequencies from low frequency patches of an image - resulting in the removal of low frequency information as well. In addition to this, the precalculated quantization matrices for the chrominance and luminance channels are specifically designed to keep the image unchanged to the human eye. Thus, while JPEG compression may remove the harmful perturbation, it is difficult to pinpoint where exactly in the frequency spectrum this occurred. Prior to our work, it was also unknown whether JPEG compression could break\footnote{Because the purpose of an unlearnable dataset is to make data useless for training, to ``break'' an unlearnable dataset means to make the data useful again -- to train a model to high test accuracy using the corrupted data.} CUDA, a fundamentally different type of poison with relatively larger perturbations. In contrast, our method includes a decomposition of an image into spatial frequencies, allowing us to determine where perturbations lie. 

\section{Motivation}


DNNs fail to generalize to the data because the perturbations are tied with the class labels \cite{wu2023onepixel, Sadasivan_2023_CVPR}. This phenomena is known as shortcut learning, where DNNs learn the simplest feature that correlates well with labels. 
In this section, we empirically verify that DNNs in fact do learn shortcuts when trained on unlearnable data. 

\subsection{Shortcut Learning}
To isolate the effect of perturbations, we must ensure the model does not learn meaningful features from the training dataset. All the unlearnable methods we analyze employ a class-wise poison, creating a tie between perturbation and class label. Suppose \(K\) is the number of classes for a given dataset, with \(\delta \in \Delta_C\) being a perturbation. We denote the test image and test label as \(x_i\) and \(y_i\) respectively. We then apply \(\delta\) to \(x_i\) in three different ways: perturbation matching test label, perturbation not matching test label, and perturbation being random. We test the poisoned model on these three sets of the test data and expect test accuracies of 100\%, 0\%, and \(1/K\%\), respectively. 
If the model learned a shortcut, the perturbation's class should determine the predicted label. We expect 100\% when the perturbation matches the test label because the perturbation's class is always the correct class label. We expect 0\% when the test label does not match the perturbation, because the perturbation's class is always incorrect. We expect 1/$K$\% with $K$ classes when the perturbation is uniformly random, because the perturbation will only match the correct label 1/$K$\% of the time on average. Table \ref{tab:filter-table} demonstrates that poisoned models indeed learn shortcuts where the presence of a perturbation determines the model predictions.

\begin{table}[]
\centering
\begin{sc}
\begin{tabular}{ccccc}
\toprule
\multicolumn{1}{l}{}                  & \multicolumn{4}{c}{Poisoned Training Data} \\
Test Image Perturbation               & CUDA \cite{Sadasivan_2023_CVPR}     & OPS \cite{wu2023onepixel}     & AR \cite{sandoval-segura2022autoregressive}       & R4 \cite{sandoval2022poisons}      \\ \midrule
\(x_i + \delta_{y_i}\)              & 93.17     & 86.33    & 100      & 99.99    \\
\(x_i + \delta_{y_{i+1}}\)          & 6.38      & 0.86     & 0        & 0.25     \\
\(x_i + \delta_{\mathcal{U}(0, K)}\) & 14.88     & 12.41    & 10.40    & 10.42   
\end{tabular}
\end{sc}
\vspace{0.2cm}
\caption{\textbf{Unlearnable datasets train models to learn perturbations.} Using poisoned models from four unlearnable datasets, we evaluate classification accuracy on a \textit{modified test set}. Each of these unlearnable datasets applies perturbations in a class-wise manner: $\delta_{y_i}$ refers to a perturbation for class $i$ (or generated by the $i^{\rm th}$ process in the case of AR Poisoning). The test dataset is modified in three ways: \(x_i + \delta_{y_i}\) refers to the perturbation matching the class of the test image, \(x_i + \delta_{y_{i+1}}\) refers to the perturbation not matching the test image, and \(x_i + \delta_{\mathcal{U}(0, K)}\) refers to the perturbation being random. Achieving high accuracy on images not seen during training only when the right perturbation is present suggests that poisoned RN-18's learn a class perturbation, as opposed to any useful features from data.}
\label{tab:filter-table}
\end{table}

\subsection{Frequency Filtering via Discrete Cosine Transform}
\label{subsection:frequency-filtering}

\begin{figure}[h]
    \centering
    \resizebox{9cm}{!}{
        \includegraphics{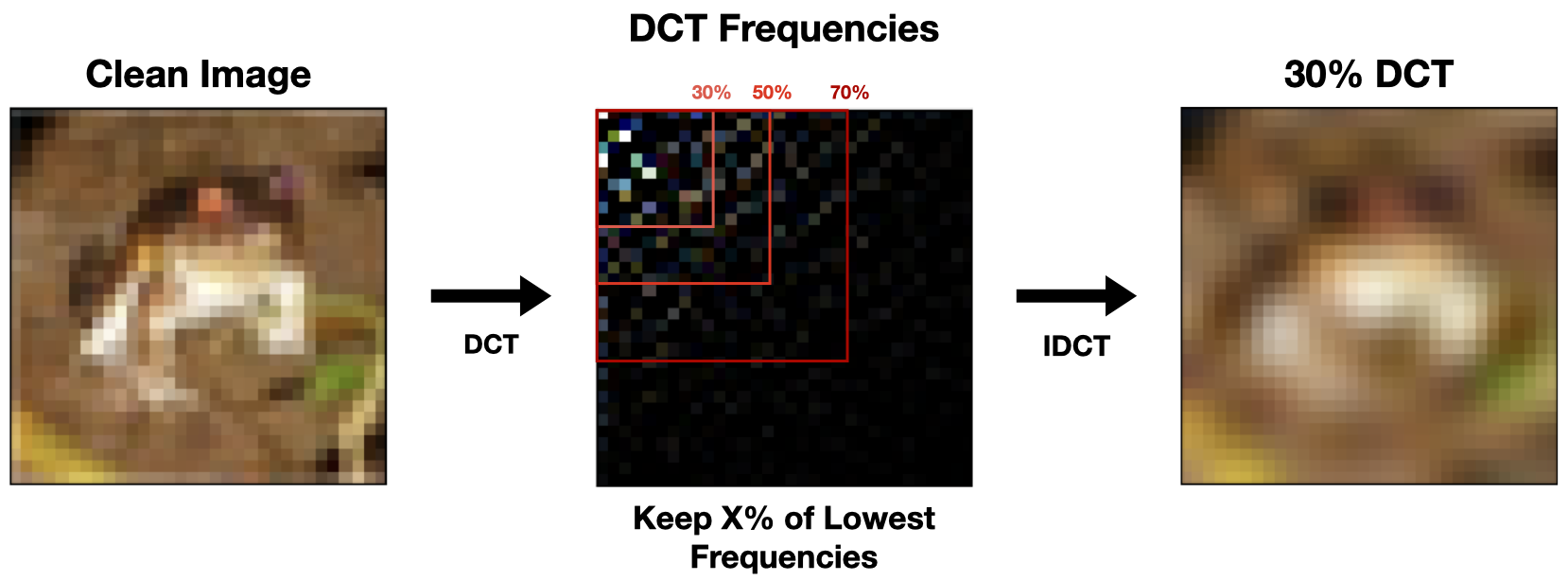}
    }
    \caption{\textbf{DCT can be used to remove exact frequency bands.} DCT converts an image into a spatial frequency representation of the same size. The coefficients represent increasing frequencies from top-left (lowest) to bottom-right (highest). We retain the lowest $X$\% of frequency coefficients by masking out the remaining higher frequencies. Then, we apply the inverse DCT (IDCT) to transform the modified frequency representation back into an image.}
    \label{fig:dctexp}
\end{figure}

\citet{liu2023image} demonstrate that JPEG compression effectively counters high-frequency perturbations, while grayscale conversion mitigates low-frequency perturbations. Although their empirical results support the efficacy of these methods, they do not delve into further analysis of how JPEG compression alters the frequencies of poisoned images. The complexity of the JPEG algorithm, which involves six discrete steps, makes it challenging to pinpoint which aspects contribute to the improvements in robustness against adversarial attacks.

Instead, we use a frequency filtering (FF) method that uses the Discrete Cosine Transform (DCT), a component of JPEG compression, to systematically remove high frequencies from images. Figure ~\ref{fig:dctexp} visualizes our method. It involves applying DCT to input images, keeping increasing percentages (10\%, 20\%, 30\%...) of the DCT coefficients, and then performing an inverse DCT to reconstruct the images. This process generates a series of images predominantly composed of lower frequency information, with higher frequency components gradually introduced as a larger fraction of coefficients is retained, resulting in images that go from blurry to sharp (refer to Figure \ref{fig:dctcomp}).

\section{Method and Experiments}
We identified shortcut learning behavior using unlearnable datasets like OPS, AR, and Regions-4. Now we discuss how we arrived at our method of learning from unlearnable datasets, and finally verify empirically that our method breaks said shortcuts. 


\subsection{Settings}

\paragraph{Datasets and Models.}
For our experiments, we train ResNet-18~\cite{he2016deep}, and three datasets: CIFAR-10~\cite{krizhevsky2009learning}, CIFAR-100~\cite{krizhevsky2009learning}, and a 100-class subset of ImageNet~\cite{russakovsky2015imagenet}, which we denote ImageNet-100. ImageNet-100 contains 129,395 train samples and 5,000 test samples. We create CUDA datasets for CIFAR-10, CIFAR-100, and ImageNet-100 using default settings from the original work: For CIFAR-10 and CIFAR-100, we use blur parameter $p_b=0.3$ and kernel size of $3$. For ImageNet-100, we use $p_b=0.06$ and kernel size of $9$. The percentage of poisoned data is $1.0$, \textit{i.e.} the whole dataset. 

\paragraph{Training Hyperparameters.}
All classifiers are trained using cross-entropy loss. For both the CIFAR-10 and CIFAR-100 datasets, we train models for 60 epochs using SGD with a learning rate of 0.1, momentum of 0.9, and weight decay of \num{5e-4}. We use a cosine annealing learning rate scheduler. We use a batch size of 512 for both, and normalize using the mean and standard deviation from each dataset. For data augmentations, we use random crops and random horizontal flips for CIFAR-10. For CIFAR-100, we additionally apply random rotations. 

For ImageNet-100, we train ResNet-18 for 100 epochs using SGD with momentum of 0.9, and weight decay of \num{5e-5}. We use a cyclic learning rate schedule which starts at $0.25$, peaks to $0.5$ at epoch $2$, and decays to $0$ by epoch $100$ (\textit{i.e.} one cycle). For data augmentations, we use random resized crops and horizontal flips as data augmentations.

\begin{figure*}[t]
    \centering
    \resizebox{13cm}{!}{
        $
        \begin{bmatrix}
            0 & -1 & 0\\
            -1 & 5 & -1\\
            0 & -1 & 0\\
        \end{bmatrix}
        \hspace{.8cm}
        \begin{bmatrix}
            0 & -0.375 & 0\\
            -0.375 & 2.5 & -0.375\\
            0 & -0.375 & 0\\
        \end{bmatrix}
        \hspace{.8cm}
        \begin{bmatrix}
            0 & -\mathcal{N}(0.375, 0.1) & 0\\
            -\mathcal{N}(0.375, 0.1) & -\mathcal{N}(2.5, 0.1) & -\mathcal{N}(0.375, 0.1)\\
            0 & -\mathcal{N}(0.375, 0.1) & 0\\
        \end{bmatrix}
        $
    }
    \caption{\textbf{Left:} A standard sharpening kernel, using a center value of 5. \textbf{Middle:} A softer sharpening kernel with a center value of 2.5. \textbf{Right:} A random sharpening kernel where values are sampled independently from a normal distribution.}
    \label{fig:kernels}
\end{figure*}

\subsection{Random Sharpening Kernels + Frequency Filtering}
\label{subsection:rsk_ff}
\paragraph{Sharpening Kernels.}
In computer vision, sharpening kernels enhance image edges by emphasizing differences in adjacent pixel values. Figure \ref{fig:kernels} (left) shows a standard sharpening kernel. We find that a softer kernel with a center value of 2.5 (Figure \ref{fig:kernels} middle) outperforms the standard kernel with a center of 5. Applying this softer kernel to CUDA CIFAR-10 training data improves RN-18 test accuracy from 23.25\% to 26.72\%, a 3.47\% increase. Despite the CUDA training data being more legible with the help of sharpening kernels, they alone cannot break the learned relationships between class-wise blur filters from CUDA and ground truth labels. Thus, we resort to randomized sharpening kernels.


\paragraph{The Random Sharpening Kernel (RSK).}
We propose to use random sharpening kernels (RSK) where we sample the center value from a normal distribution, then sample the neighboring values using the mean from the sampled center value (see Figure~\ref{fig:kernels} right). RSK samples its center value from a normal distribution with a mean of 2.5. From the previous section with SSK, we have a test accuracy of 26.72\% on CIFAR-10. After applying the RSK to the CUDA training data, we observe a 3.22\% improvement to 29.94\% test accuracy on CIFAR-10. In Table \ref{table:transforms_table}, we also include data for CIFAR-100 and ImageNet-100. An example of an RSK is in Figure \ref{fig:kernels}(c). 

\paragraph{Frequency Filtering via DCT.}

Along with RSK, we consider FF (refer to Section \ref{subsection:frequency-filtering}) as a transform. We test this method on CUDA, and CUDA preprocessed by SSK and RSK. Our results are shown in Table \ref{tab:filter-table}. We observe that keeping the lowest $30$ to $40\%$ of frequencies is best for CIFAR-10 and CIFAR-100, while a smaller percentage of $5\%$ is best for ImageNet-100. This difference is due to ImageNet images being of larger size ($224 \times 224$). However, note that $5\%$ of an ImageNet image's spatial frequencies is equivalent to $34\%$ of the spatial frequencies in a CIFAR image. \footnote{The lowest $5\%$ of spatial frequencies in an ImageNet image is represented by a $11 \times 11$ DCT coefficients matrix, which is $\frac{11}{32} \approx 34$ percent of a CIFAR image's spatial frequencies.}

\paragraph{Putting it all together.}
As seen in Table \ref{table:transforms_table}, FF and RSK do not perform well by themselves against CUDA, but they excel when used together. After applying RSK with FF (30), we observe a test accuracy jump from 29.94\% to 78.53\% on CIFAR-10, an improvement of $48.59\%$ over relying on RSK alone. We find that using an RSK with center 2.5 and FF with 30\% of lowest frequencies kept is the best way to learn from CUDA. 

\begin{table}[h]
\centering
\begin{sc}
\resizebox{11cm}{!}{
\begin{tabular}{lcccc}
\toprule

\multicolumn{1}{c}{} & CIFAR-10       & CIFAR-100      & ImageNet-100   \\ \midrule
Clean                         & 94.34          & 74.41          & 81.76          \\ \midrule
CUDA                          & 23.25          & 17.33          & 2.68           \\
CUDA + SSK                    & 26.48          & 23.82          & 5.04            \\
CUDA + RSK                    & 29.01          & 27.27          & 6.34            \\
CUDA + FF                     & 37.48 (10)     & 21.06 (30)     & 40.84 (5)      \\ \midrule
CUDA + SSK + FF               & 72.60 (30)     & 50.87 (30)     & 42.58 (5)            \\
CUDA + RSK + FF               & \textbf{78.53 (30)} & \textbf{53.33 (40)} & \textbf{43.08 (5)}


\end{tabular}
}
\end{sc}
\vspace{0.2cm}
\caption{\textbf{A breakdown of the transforms we use against CUDA.} It shows the test accuracies of the different combinations of transforms we use (refer to Section \ref{subsection:rsk_ff} for a breakdown). The parentheses next to results that use FF refers to the \% of lowest frequencies kept. See Appendix \ref{appendix:ff-analysis} for an analysis on different percentages kept. }
\label{table:transforms_table}
\end{table}


 
\section{Conclusion}

As protection against data scraping and copyright infringement, unlearnable datasets must be robust against adversarial training and other attacks. We identified that shortcut learning is a technique used by unlearnable datasets to fool DNNs. Based on the fact that CUDA is a class-wise blur, we designed a random sharpening kernel that breaks the relationship between the filters and their corresponding labels. We also used the intuition of JPEG compression being capable of removing perturbative information, and designed a method to filter out specific frequencies using the Discrete Cosine Transform. We found that we were able to boost test accuracy from CUDA-convolved CIFAR-10, CIFAR-100, and ImageNet-100 by 55.25\%, 36\%, and 40.4\%, respectively. From our results, we find that unlearnable datasets can be exploited by simple image transformations, making them not as unlearnable as once thought. 

\section*{Acknowledgments}
This work was made possible by National Science Foundation (NSF) grant \#2213335. Pedro is supported by a National Defense Science and Engineering Graduate (NDSEG) Fellowship.

\newpage
{\small
\bibliographystyle{plainnat}
\bibliography{references}

\begin{thebibliography}{21}
\providecommand{\natexlab}[1]{#1}
\providecommand{\url}[1]{\texttt{#1}}
\expandafter\ifx\csname urlstyle\endcsname\relax
  \providecommand{\doi}[1]{doi: #1}\else
  \providecommand{\doi}{doi: \begingroup \urlstyle{rm}\Url}\fi

\bibitem[Carlini et~al.(2024)Carlini, Jagielski, Choquette-Choo, Paleka, Pearce, Anderson, Terzis, Thomas, and Tramèr]{carlini2023poisoning}
N.~Carlini, M.~Jagielski, C.~Choquette-Choo, D.~Paleka, W.~Pearce, H.~Anderson, A.~Terzis, K.~Thomas, and F.~Tramèr.
\newblock Poisoning web-scale training datasets is practical.
\newblock In \emph{2024 IEEE Symposium on Security and Privacy (SP)}, pages 176--176, Los Alamitos, CA, USA, may 2024. IEEE Computer Society.
\newblock \doi{10.1109/SP54263.2024.00094}.
\newblock URL \url{https://doi.ieeecomputersociety.org/10.1109/SP54263.2024.00094}.

\bibitem[Chen et~al.(2017)Chen, Liu, Li, Lu, and Song]{chen2017targeted}
Xinyun Chen, Chang Liu, Bo~Li, Kimberly Lu, and Dawn Song.
\newblock Targeted backdoor attacks on deep learning systems using data poisoning.
\newblock \emph{arXiv preprint arXiv:1712.05526}, 2017.

\bibitem[Cohen et~al.(2019)Cohen, Rosenfeld, and Kolter]{cohen2019certified}
Jeremy Cohen, Elan Rosenfeld, and Zico Kolter.
\newblock Certified adversarial robustness via randomized smoothing.
\newblock In \emph{international conference on machine learning}, pages 1310--1320. PMLR, 2019.

\bibitem[Das et~al.(2017)Das, Shanbhogue, Chen, Hohman, Chen, Kounavis, and Chau]{das2017keeping}
Nilaksh Das, Madhuri Shanbhogue, Shang-Tse Chen, Fred Hohman, Li~Chen, Michael~E Kounavis, and Duen~Horng Chau.
\newblock Keeping the bad guys out: Protecting and vaccinating deep learning with jpeg compression.
\newblock \emph{arXiv preprint arXiv:1705.02900}, 2017.

\bibitem[Desai and Johnson(2021)]{desai2021virtex}
Karan Desai and Justin Johnson.
\newblock Virtex: Learning visual representations from textual annotations.
\newblock In \emph{Proceedings of the IEEE/CVF conference on computer vision and pattern recognition}, pages 11162--11173, 2021.

\bibitem[Fowl et~al.(2021)Fowl, Goldblum, Chiang, Geiping, Czaja, and Goldstein]{fowl2021adversarialpoison}
Liam Fowl, Micah Goldblum, Ping-yeh Chiang, Jonas Geiping, Wojciech Czaja, and Tom Goldstein.
\newblock Adversarial examples make strong poisons.
\newblock In M.~Ranzato, A.~Beygelzimer, Y.~Dauphin, P.S. Liang, and J.~Wortman Vaughan, editors, \emph{Advances in Neural Information Processing Systems}, volume~34, pages 30339--30351. Curran Associates, Inc., 2021.
\newblock URL \url{https://proceedings.neurips.cc/paper/2021/file/fe87435d12ef7642af67d9bc82a8b3cd-Paper.pdf}.

\bibitem[Geirhos et~al.(2020)Geirhos, Jacobsen, Michaelis, Zemel, Brendel, Bethge, and Wichmann]{geirhos2020shortcut}
Robert Geirhos, J{\"o}rn-Henrik Jacobsen, Claudio Michaelis, Richard Zemel, Wieland Brendel, Matthias Bethge, and Felix~A Wichmann.
\newblock Shortcut learning in deep neural networks.
\newblock \emph{Nature Machine Intelligence}, 2\penalty0 (11):\penalty0 665--673, 2020.

\bibitem[He et~al.(2016)He, Zhang, Ren, and Sun]{he2016deep}
Kaiming He, Xiangyu Zhang, Shaoqing Ren, and Jian Sun.
\newblock Deep residual learning for image recognition.
\newblock In \emph{Proceedings of the IEEE conference on computer vision and pattern recognition}, pages 770--778, 2016.

\bibitem[Huang et~al.(2021)Huang, Ma, Erfani, Bailey, and Wang]{huang2021unlearnable}
Hanxun Huang, Xingjun Ma, Sarah~Monazam Erfani, James Bailey, and Yisen Wang.
\newblock Unlearnable examples: Making personal data unexploitable.
\newblock In \emph{International Conference on Learning Representations}, 2021.
\newblock URL \url{https://openreview.net/forum?id=iAmZUo0DxC0}.

\bibitem[Krizhevsky et~al.(2009)Krizhevsky, Hinton, et~al.]{krizhevsky2009learning}
Alex Krizhevsky, Geoffrey Hinton, et~al.
\newblock Learning multiple layers of features from tiny images.
\newblock 2009.

\bibitem[Liu et~al.(2023)Liu, Zhao, and Larson]{liu2023image}
Zhuoran Liu, Zhengyu Zhao, and Martha Larson.
\newblock Image shortcut squeezing: Countering perturbative availability poisons with compression.
\newblock In \emph{International conference on machine learning}, pages 22473--22487. PMLR, 2023.

\bibitem[Nasr et~al.(2018)Nasr, Shokri, and Houmansadr]{nasr2018comprehensive}
Milad Nasr, Reza Shokri, and Amir Houmansadr.
\newblock Comprehensive privacy analysis of deep learning.
\newblock In \emph{Proceedings of the 2019 IEEE Symposium on Security and Privacy (SP)}, volume 2018, pages 1--15, 2018.

\bibitem[Russakovsky et~al.(2015)Russakovsky, Deng, Su, Krause, Satheesh, Ma, Huang, Karpathy, Khosla, Bernstein, et~al.]{russakovsky2015imagenet}
Olga Russakovsky, Jia Deng, Hao Su, Jonathan Krause, Sanjeev Satheesh, Sean Ma, Zhiheng Huang, Andrej Karpathy, Aditya Khosla, Michael Bernstein, et~al.
\newblock Imagenet large scale visual recognition challenge.
\newblock \emph{International journal of computer vision}, 115:\penalty0 211--252, 2015.

\bibitem[Sadasivan et~al.(2023)Sadasivan, Soltanolkotabi, and Feizi]{Sadasivan_2023_CVPR}
Vinu~Sankar Sadasivan, Mahdi Soltanolkotabi, and Soheil Feizi.
\newblock Cuda: Convolution-based unlearnable datasets.
\newblock In \emph{Proceedings of the IEEE/CVF Conference on Computer Vision and Pattern Recognition (CVPR)}, pages 3862--3871, June 2023.

\bibitem[Sandoval-Segura et~al.(2022{\natexlab{a}})Sandoval-Segura, Singla, Fowl, Geiping, Goldblum, Jacobs, and Goldstein]{sandoval2022poisons}
Pedro Sandoval-Segura, Vasu Singla, Liam Fowl, Jonas Geiping, Micah Goldblum, David Jacobs, and Tom Goldstein.
\newblock Poisons that are learned faster are more effective.
\newblock In \emph{Proceedings of the IEEE/CVF Conference on Computer Vision and Pattern Recognition (CVPR) Workshops}, pages 198--205, June 2022{\natexlab{a}}.

\bibitem[Sandoval-Segura et~al.(2022{\natexlab{b}})Sandoval-Segura, Singla, Geiping, Goldblum, Goldstein, and Jacobs]{sandoval-segura2022autoregressive}
Pedro Sandoval-Segura, Vasu Singla, Jonas Geiping, Micah Goldblum, Tom Goldstein, and David Jacobs.
\newblock Autoregressive perturbations for data poisoning.
\newblock \emph{Advances in Neural Information Processing Systems}, 35:\penalty0 27374--27386, 2022{\natexlab{b}}.

\bibitem[Shaopeng et~al.(2022)Shaopeng, Fengxiang, Yang, Li, and Dacheng]{fu2022robust}
Fu~Shaopeng, He~Fengxiang, Liu Yang, Shen Li, and Tao Dacheng.
\newblock Robust unlearnable examples: Protecting data privacy against adversarial learning.
\newblock In \emph{International Conference on Learning Representations}, 2022.
\newblock URL \url{https://openreview.net/forum?id=baUQQPwQiAg}.

\bibitem[Wang et~al.(2020)Wang, Wu, Huang, and Xing]{wang2020high}
Haohan Wang, Xindi Wu, Zeyi Huang, and Eric~P Xing.
\newblock High-frequency component helps explain the generalization of convolutional neural networks.
\newblock In \emph{Proceedings of the IEEE/CVF conference on computer vision and pattern recognition}, pages 8684--8694, 2020.

\bibitem[Wu et~al.(2023)Wu, Chen, Xie, and Huang]{wu2023onepixel}
Shutong Wu, Sizhe Chen, Cihang Xie, and Xiaolin Huang.
\newblock One-pixel shortcut: On the learning preference of deep neural networks.
\newblock In \emph{The Eleventh International Conference on Learning Representations}, 2023.
\newblock URL \url{https://openreview.net/forum?id=p7G8t5FVn2h}.

\bibitem[Yu et~al.(2022)Yu, Zhang, Chen, Yin, and Liu]{yu2022availability}
Da~Yu, Huishuai Zhang, Wei Chen, Jian Yin, and Tie-Yan Liu.
\newblock Availability attacks create shortcuts.
\newblock In \emph{Proceedings of the 28th ACM SIGKDD Conference on Knowledge Discovery and Data Mining}, pages 2367--2376, 2022.

\bibitem[Yuan and Wu(2021)]{yuan21ntga}
Chia-Hung Yuan and Shan-Hung Wu.
\newblock Neural tangent generalization attacks.
\newblock In Marina Meila and Tong Zhang, editors, \emph{Proceedings of the 38th International Conference on Machine Learning}, volume 139 of \emph{Proceedings of Machine Learning Research}, pages 12230--12240. PMLR, 18--24 Jul 2021.
\newblock URL \url{https://proceedings.mlr.press/v139/yuan21b.html}.

\end{thebibliography}
}

\newpage
\appendix

\section{Appendix}

\subsection{Analysis on Frequency Filtering}
\label{appendix:ff-analysis}

In this section, we analyze how datasets and different combinations of transforms behave under multiple percentages of kept DCT frequencies. 

\begin{table}[h]
\centering

\begin{tabular}{lcccccccccc}
\toprule
           & \multicolumn{10}{c}{Kept DCT Frequency Percentage}                                                       \\
           & 10\%           & 20\%  & 30\%           & 40\%  & 50\%  & 60\%  & 70\%  & 80\%  & 90\%  & 100\%          \\ \midrule
Clean      & 40.74          & 37.48 & 46.55          & 31.73 & 44.85 & 91.40 & 92.16 & 92.57 & 93.18 & \textbf{93.66} \\ \midrule
CUDA       & \textbf{37.48} & 36.83 & 21.09          & 20.36 & 20.20 & 17.51 & 20.35 & 21.37 & 21.95 & 21.04          \\
SSK        & 45.26          & 71.06 & \textbf{72.60} & 58.09 & 49.22 & 45.07 & 41.61 & 38.12 & 30.19 & 26.48          \\
RSK        & 46.55          & 73.72 & \textbf{78.64} & 68.10 & 60.35 & 53.86 & 48.60 & 41.23 & 35.95 & 29.01          
\end{tabular}

\vspace{0.2cm}
\caption{\textbf{RSK is better than SSK.} This table displays test accuracies of ResNet-18 trained on CIFAR-10. CUDA refers to only CUDA convolved images. RSK refers to CUDA images convolved with RSK, and SSK refers to CUDA images convolved with SSK. It is observed that RSK has better performance than SSK on all frequency bands. 
}
\label{tab:ff-cifar10}
\end{table}
The baseline, clean CIFAR-10, performs as expected. As we allow the model to train on an increasing amount of higher frequency bands to the model, the test accuracy improves as high frequency components are useful for classification \cite{wang2020high}. For the next three transforms, as seen in Table~\ref{tab:ff-cifar10}, we can see a trend: they perform best when most of the high frequencies are masked out; in other words, mostly blurry. This is interesting because from previous works, we know that high frequency components help DNNs generalize better. However, Table~\ref{tab:ff-cifar10} shows that high frequency components can in fact ``distract" the model from learning. We posit that sharpening kernels bring back crucial information needed for DNNs to generalize, and removing high frequencies works as an effective countermeasure to break the relation between the class-wise filters from CUDA and their corresponding labels. 

\begin{figure}[h]
    \centering
    \resizebox{13.5cm}{!}{
        \includegraphics{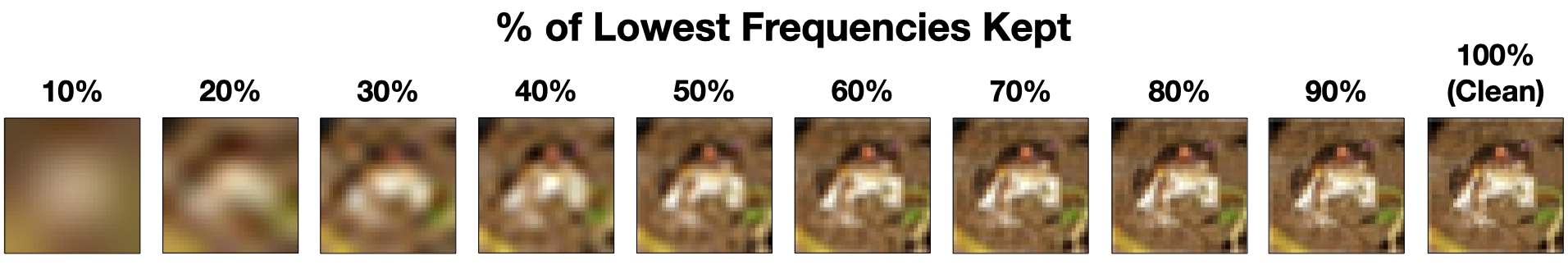}
    }
    \caption{\textbf{Image reconstruction using different percentages of frequencies kept.} On the very left, we retain the lowest 10\% of DCT frequencies. We progressively preserve higher frequencies, resulting in a series of blurry to clear images. These images are constructed using an image from the clean dataset. }
    \label{fig:dctcomp}
\end{figure}

\begin{table}[h]
\centering

\begin{tabular}{lcccccccccc}
\toprule
           & \multicolumn{10}{c}{Kept DCT Frequency Percentage}                                                       \\
           & 10\%  & 20\%  & 30\%           & 40\%  & 50\%  & 60\%  & 70\%  & 80\%  & 90\%  & 100\%          \\ \midrule
Clean      & 17.14 & 39.53 & 54.16          & 61.05 & 65.74 & 69.09 & 70.14 & 71.19 & 72.29 & \textbf{72.80} \\ \midrule
CUDA       & 13.50 & 21.04 & \textbf{21.06} & 13.76 & 12.74 & 11.03 & 11.47 & 13.00 & 12.65 & 15.35          \\
SSK        & 19.10 & 40.54 & \textbf{50.87} & 50.57 & 47.12 & 44.30 & 39.88 & 29.85 & 24.44 & 23.82          \\
RSK        & 19.90 & 42.47 & 52.58 & \textbf{53.33} & 52.84 & 49.55 & 46.08 & 35.44 & 29.23 & 27.27        
\end{tabular}

\vspace{0.2cm}
\caption{\textbf{Test accuracies of ResNet-18 trained on CIFAR-100.} We test our sharpening kernels, SSK and RSK, with DCT filtering (FF) on different percentages of kept DCT frequencies. We find that percentages of 30\% to 40\% work the best, similar to CIFAR-10. }

\label{tab:ff-cifar100}
\end{table}

\begin{table}[H]
\centering

\begin{tabular}{lccccccccccc}
\toprule
           & \multicolumn{10}{c}{Kept DCT Frequency Percentage}                                              \\
           & 5\%   & 10\%   & 20\%  & 30\%  & 40\%  & 50\%  & 60\%  & 70\%  & 80\%  & 90\%  & 100\%          \\ \midrule
CUDA       & \textbf{40.84} & 38.76 & 4.14  & 2.86  & 3.28  & 2.02  & 2.76  & 2.40  & 2.88  & 2.96 & 2.74    \\
SSK        & \textbf{42.58} & 39.04 & 4.46  & 5.96  & 5.42  & 7.08  & 5.90  & 4.72  & 4.62  & 5.98 & 5.38    \\
RSK        & \textbf{43.08} & 39.64 & 8.76  & 7.32  & 5.62  & 7.44  & 6.26  & 6.28  & 8.54  & 6.50 & 7.52                     
\end{tabular}

\vspace{0.2cm}
\caption{\textbf{Test accuracies of ResNet-18 trained on ImageNet-100.} We test our sharpening kernels, SSK and RSK, with DCT filtering (FF) on different percentages of kept DCT frequencies. We find that a percentage of 5\% works the best. The lowest 5\% of spatial frequencies in ImageNet images is represented by an 11 $\times$ 11 DCT coefficient matrix, which is $\frac{11}{32} \approx 34$ percent of a CIFAR image's spatial frequencies - it matches up with the percentages where highest accuracies were achieved with CIFAR-10 and CIFAR-100 (30\% and 40\% each).}
\label{tab:ff-imagenet100}
\end{table}

\begin{table}[h]
\centering
\begin{tabular}{lllll}

\toprule
         & CUDA\cite{Sadasivan_2023_CVPR}       & OPS\cite{wu2023onepixel}        & AR\cite{sandoval-segura2022autoregressive}        & R4\cite{sandoval2022poisons}        \\ \midrule
Baseline & 23.25      & 31.05      & 18.41      & 14.04      \\
RSK + FF & 78.53 (30) & 87.03 (40) & 67.23 (20) & 23.17 (10)
\end{tabular}
\vspace{0.2cm}
\caption{\textbf{Test accuracies of ResNet-18 trained on different sets of unlearnable datasets on CIFAR-10.} We tested various unlearnable datasets against our best method, RSK + FF. We find that it performs even better on OPS than CUDA, but performs poorly on Autoregressive Poisons and Regions-4. The parentheses notes the lowest kept percentage of DCT frequencies, and we reported the highest accuracy after testing from 10\% to 100\%.}
\label{tab:ff-uds}

\end{table}

\end{document}